\documentclass[sigconf]{acmart}
\usepackage{graphicx}
\AtBeginDocument{%
  \providecommand\BibTeX{{%
    \normalfont B\kern-0.5em{\scshape i\kern-0.25em b}\kern-0.8em\TeX}}}

\setcopyright{acmcopyright}

\begin{document}

\title{Wound and Episode Level Readmission Risk or Weeks to Readmit: Why do patients get readmitted? How Long Does It Take for a patient to get readmitted?}


\author{Subba Reddy Oota}
\email{soota@woundtech.net}
\affiliation{%
  \institution{Woundtech Innovative Healthcare Solutions}
}

\author{Nafisur Rahman}
\email{nrahman@woundtech.net}
\affiliation{%
  \institution{Woundtech Innovative Healthcare Solutions}
  }

\author{Shahid Saleem Mohammed}
\email{shmohammed@woundtech.net}
\affiliation{%
  \institution{Woundtech Innovative Healthcare Solutions}
}

\author{Jeffrey Galitz}
\affiliation{%
\institution{Woundtech Innovative Healthcare Solutions}
 }

\author{Ming Liu}
\affiliation{%
  \institution{Woundtech Innovative Healthcare Solutions}
  }

\renewcommand{\shortauthors}{Subba and Nafisur, et al.}

\begin{abstract}
 The Affordable care Act of 2010 had introduced Readmission reduction program in 2012 to reduce avoidable re-admissions to control rising healthcare costs. Wound care impacts 15\%~\cite{nussbaum2018economic} of medicare beneficiaries making it one of the major contributors of medicare health care cost.Health plans have been exploring proactive health care services that can focus on preventing wound recurrences and re-admissions to control the wound care costs. With rising costs of Wound care industry, it has become of paramount importance to reduce wound recurrences \& patient re-admissions. What factors are responsible for a Wound to recur which ultimately lead to hospitalization or re-admission? Is there a way to identify the patients at risk of re-admission before the occurrence using data driven analysis? Patient re-admission risk management has become critical for patients suffering from chronic wounds such as diabetic ulcers, pressure ulcers, and vascular ulcers. Understanding the risk \& the factors that cause patient readmission can help care providers and patients avoid wound recurrences. Our work focuses on identifying patients who are at high risk of re-admission \& determining the time period with in which a patient might get re-admitted. Frequent re-admissions add financial  stress to the patient \& Health plan and deteriorate the quality of life of the patient.Having this information can allow a provider to set up preventive measures that can delay, if not prevent, patients' re-admission.
 
 On a combined wound \& episode-level data set of patient's wound care information, our extended autoprognosis achieves a recall of 92 and a precision of 92 for the predicting a patient's re-admission risk. 
For new patient class, precision and recall are as high as 91 and 98, respectively. 
We are also able to predict the amount of time (in weeks) it might take after a patient's discharge event for a re-admission event to occur through our model with a mean absolute error of 2.3 weeks.

\end{abstract}

\keywords{Patient's re-admission risk, Auto prognosis, Health care, Wound care, Chronic Wound management, Re-admission prevention, Cost Control, Machine Learning}

\maketitle

\section{Introduction}
Nearly 3.3 million patients were readmitted to the hospital within 30 days of being discharged in the United States as per the Agency of Healthcare Research in the year 2011~\cite{briefing2014ahrq}.
Also, over \$41 billion were spent due to patient re-admissions in 2011~\cite{briefing2014ahrq}.
For wound ulcer specific re-admissions, over \$250 million were spent on re-admissions that occurred due to diabetic wounds, and more than \$11 billion were spent on pressure ulcer related re-admissions ~\footnote{\url{https://www.ahrq.gov/sites/default/files/publications/files/putoolkit.pdf}}.
Patient re-admissions can cause a significant increase in cost and also lead to federal fines on hospitals for poor clinical outcomes. It becomes important for Wound care providers to focus on the re-admission problem \& determine the risk \& cause of the re-admission. Some of the critical questions that we try to address in our work are: (1) What factors drive wound-related re-admissions? , (2) Are patients returning with new wounds or with same wounds to the care?, (3) What is the risk of an existing wound to recur in future?, (4) How much impact does patient's non-compliance in matters such as wound dressing, diet, medication, compression, and exercise have on patient's re-admission risk? , (5) what is the overall re-admission risk for a patient provided their wound history \& non-compliance history? (6) In how many weeks, can we expect a patient to end up in the hospital due to wound-related problems? ~\cite{baillie2013readmission}.

\begin{table*}[t]%
\centering
\small
\caption{Statistics for the patient's readmission risk attributes in our dataset}
\label{tab:dataStats}
\begin{tabular}{|c|p{0.3\linewidth}|c|c|p{0.3\linewidth}|}
\hline

Dataset&Attributes&\#Instances& \#Classes&
    Top classes
\\
\hline
Wound-Level& WoundStatus, PatientDischargeStatus, PalliativeCare, Wound/Ulcer Type, Wound Location, Wound Stage, DaysinTXforWounds, AvgPainLevelforWound, VisitsforWound, DaysinTXforPatients  &90328&2& WoundRecurrence: (Recurring Wound (24886), New Wound (65442)), 
Patient Category: ( Re-AdmittedPatient (46620), NewPatient
(43708))\\
\hline
Episode-Level & WoundsforEpisode, ChronicWoundsforEpisode, \#AvgDaysinTXforwounds, AvgPainLevelforEpisode, LowerExtremityWoundsforEpisode, AVGTemperature, Diabetes, Anemia, EndStageRenalDiseasewithdialysis, VenousInsuffiency, ChronicObstructivePulmonaryDisease, AtheroscleroticHeartDisease, CoronaryArteryDisease, Smoking, Edema, PeripheralArterialDisease(PVD), EndStageRenalDiseasewithoutdialysis(CKD), Hypertension, CongestiveHeartFailure, Obesity, WeightGain, MarkedWeightChange &45261&2&Patient Category: (Re-AdmittedPatient (21521), NewPatient
(23930))\\
\hline
\multicolumn{5}{|c|}{EpisodeNumber, PtAge, NonComplianceWoundVisitsRate, NonComplianceDietRate,NonComplianceOffLoadRate
} \\
\multicolumn{5}{|c|}{NonComplianceExerciseRate, NonComplianceMedicationRate, NonComplianceLimbRate, 
NonComplianceCompressionRate,} \\
\multicolumn{5}{|c|}{NonComplianceDressingRate
NonComplianceSmokingRate, NonComplianceHBOVisitsRate} \\
\hline

\end{tabular}
\vspace{0.1cm}

\end{table*}

To conduct our research work, we have collected a data set of over 20,000 patients who were managed by a Wound care provider.Using the data engineering techniques we created 2 data sets from the initial data set which represented patient data at two different granularity's.  First data set is a representation of a patient's wound care episode with the wound care provider; a patient will have more than one episode if they were re-admitted back to our Wound care provider's facility. Second data set represents all wounds that were treated by the Wound care provider for different patients; a wound can have multiple records if it had recurred in the past. Our wound care provider had implemented a process through their Electronic Health Record (EHR) system to identify a recurring wound from a new wound which helped us in determining a new wound from a recurring wound. This is an important piece of information for our work as we are trying to determine the impact of recurring wounds when a patient re-admits. Almost 50\% of our re-admit patients returned to care due to a recurring wound making wound recurrence a major factor for re-admissions. The remaining patient's returned to care due to new wounds which were a result of multiple factors such as old age, existing co-morbid conditions \& non-compliance.

After completing the data engineering task, we had to address the problem of testing the completeness of the data. In order to ensure that we are correctly categorizing patients as readmit or Non-readmit we had to ensure that the patient who was discharged from our wound care provider's service did not end up into another Wound care provider's facility for care. In order to tackle this challenge, we limited our analysis to patients who had been discharged at least 2 months ago \& had responded to the patient engagement program, a patient outreach program to check the status of patient's Wound after their discharge. Also, all the patients who joined the services of our Wound care provider were asked to provide Wound acquired date to keep track of Wound age \& its recurrence status. We observed that among these patients 33\% patients had history of re-admission \& were among those with highest risk of re-admission. 

Our data set mainly consisted of Patient demographics data, Patient Episode facts, Patient's existing Co-morbid conditions information. Patients' attributes such as, Age, BMI, Location, Braden score can play a crucial role in determining patients' re-admission risk. Apart from this, co-morbid conditions \& non-compliance data can help us assess an individuals risk of developing new wounds. Traditionally, a patients' re-admission risk has been assessed manually by the clinicians \& care monitors by studying the entire clinical history of patient. Some of the challenges of this approach are (i) Time consuming process (ii) Lack of consistency among assessors in their assessment process when the group of patients that require assessment is large (iii) The reliability on these assessments for Clinical decision making is limited. Automating the patients' risk of re-admission can overcome most of these limitations and help the clinicians to make effective and informative decisions.

In this paper, our first goal is to build a model to predict the patient’s risk of readmission. 
To achieve the first goal, our proposed system, mainly depends on three data sets such as, (i) would-level information, (ii) episode-level information, and (iii) combination of wound and episode level information .  
Further, for Wounds that possess high risk of recurrence, our second goal is to predict the time period during which a readmission event, i.e weeks from the date of discharge, can be expected. We are interested in extracting information from the clinical records using machine learning models and AutoPrognosis~\cite{alaa2018autoprognosis}. 
Often machine learning models have strong predictive power but our main focus is to build a system for automating the design of predictive modeling pipelines tailored for clinical prognosis. A model is more useful as a clinical tool if the physician understands the features underlying its predictions. 

The main contributions of this paper are as follows.
\begin{itemize}
    \item We formulate the problem as a two-stage automation method for re-admission risk and number of weeks to readmit prediction.
    \item In the first stage, we analyze the wound-level and episode-level data sets, feature analysis for risk of readmission, and categorize the patient as New Patient or Re-admit.
    \item In the second stage, we build a number of weeks to readmit model, which uses the similar features from stage 1.
    \item We integrate the two successful machine learning models LightGBM \& CatBoost into AutoPrognosis Framework.
\end{itemize}

\section{Related Work}
Patient readmission risk models have become effective tools in clinical decision making and provide several benefits to both health care providers and patients~\cite{zheng2015predictive, eby2015predictors}.
The earlier works in the literature focused on analyzing the readmission risk from various patient data sets with different ulcer types such as, diabetic~\cite{briefing2014ahrq,silverstein2008risk}, pressure ulcers~\cite{dzwierzynski1998improvement}.
The patient readmission risk prediction models were built with an aim of identifying the patients with high risk of hospital re-admissions using direct specific interventions such as, demographic details of a patient, clinical procedure-related, and diagnostic-related features for patients above 65 years of age. These models have shown to effectively diminish the readmission
rates for patients after hospital discharge~\cite{kripalani2014reducing,urma2017interventions,leppin2014preventing}. 
However, these studies discriminate poorly on re-admissions due to non-availability of patient's demographic details, medication reconciliation, and patient's education details, etc,.

\noindent{\textbf{Machine Learning Models for Re-Admission Risk}}
Motivated by the immense success of machine learning \& deep learning models in AI, there has been a lot of focus on applying machine learning \& deep learning on electronic medical records. State-of-the-art AI solutions that have demonstrated  high performance in diagnosing \& detecting diseases, risk prediction~\cite{futoma2015comparison}, and patient sub-typing~\cite{baytas2017patient,che2017rnn} have been a source of motivation for our research work.
All the existing works on patient's re-admission risk have used simple machine learning classifiers for risk prediction such as, logistic regression, naive-bayes, and SVM models~\cite{ross2008statistical,choudhury2018evaluating,zheng2015predictive}.
Moreover, the model predictions results vary from hospital to hospital, data to data.
Also, these studies use traditional feature engineering methods when handling of categorical and numerical variables, and each model requires separate hyper-parameter tuning.
To overcome the above limitations, recently, an automated framework has been developed, known
as DeepSurv (deep Cox proportional hazards neural network)~\cite{katzman2018deepsurv}.
Some other existing survival models that dealt with healthcare applications include deep active analysis, deep recurrent analysis, deep integrative analysis.
Recently, a new framework (AutoPrognosis)~\cite{alaa2018autoprognosis} was developed to automate the design of predictive modeling pipelines tailored for clinical prognosis outperforms the earlier state-of-the-art models.

\section{Feature Analysis}
In this section, we have offered a detailed analysis of features at wound-level and episode-level data sets.
There are common attributes present in both the data sets, including, admission date, patient age, episode number reported in bottom row of the Table~\ref{tab:dataStats}.

\subsection{Wound-level Feature Analysis}
Table~\ref{tab:dataStats} first row showcases the attributes used in predicting risk of readmission due to wound level factors.
Out of the 23 wound-level attributes, we have discussed some of the important features such as wound type, wound location, wound stage as follows.

\begin{figure}[t]
\centering
\includegraphics[width=\linewidth]{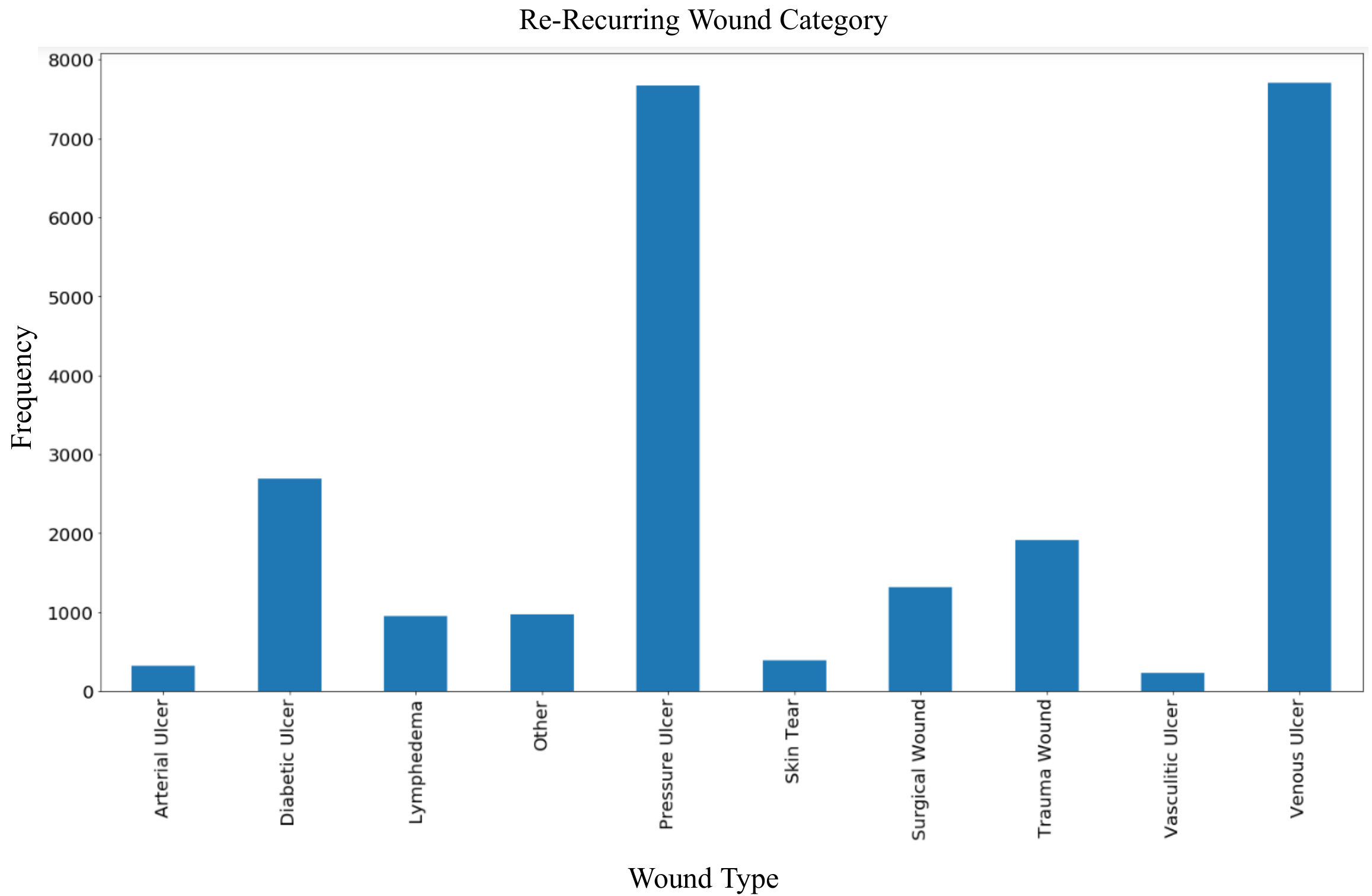}
    \caption{Wound Type Histogram in Recurring Wound Class}
    \label{fig:typehist}
\end{figure}

\noindent{\textbf{Wound Type}} Wound type attributes represent the Wound category diagnosis offered by the Physician during the initial encounter. Figure~\ref{fig:typehist} shows the frequency of recurrence of each Wound type present in our data set. Based on the observation from the Figure, one can conclude that patient's with chronic wounds such as pressure, venous \& diabetic ulcers are at high risk of re-admission.

\begin{figure}[t]
\centering
\includegraphics[width=\linewidth]{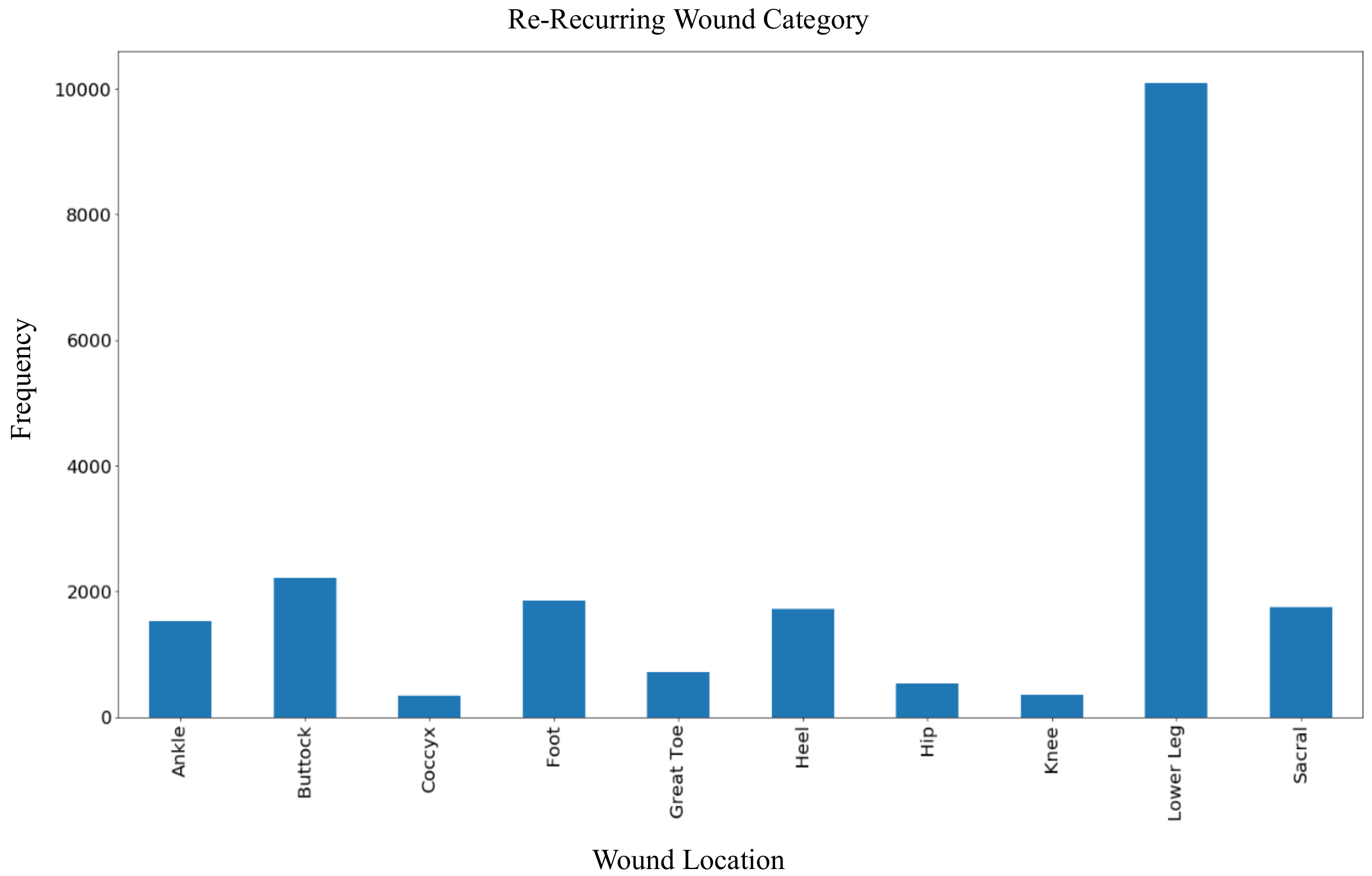}
    \caption{Wound Location Histogram in Re-Recurring Wound Class}
    \label{fig:locationhist}
\end{figure}

\noindent{\textbf{Wound Location}}
Like wound type, we observed from the Figure~\ref{fig:locationhist} that patients with lower extremity wounds have higher risk of re-admission. Also, it is common for old age patients to injure their lower extremity wounds frequently due to their deteriorating locomotive \& cognitive capabilities.

\begin{figure}[t]
\centering
\includegraphics[width=\linewidth]{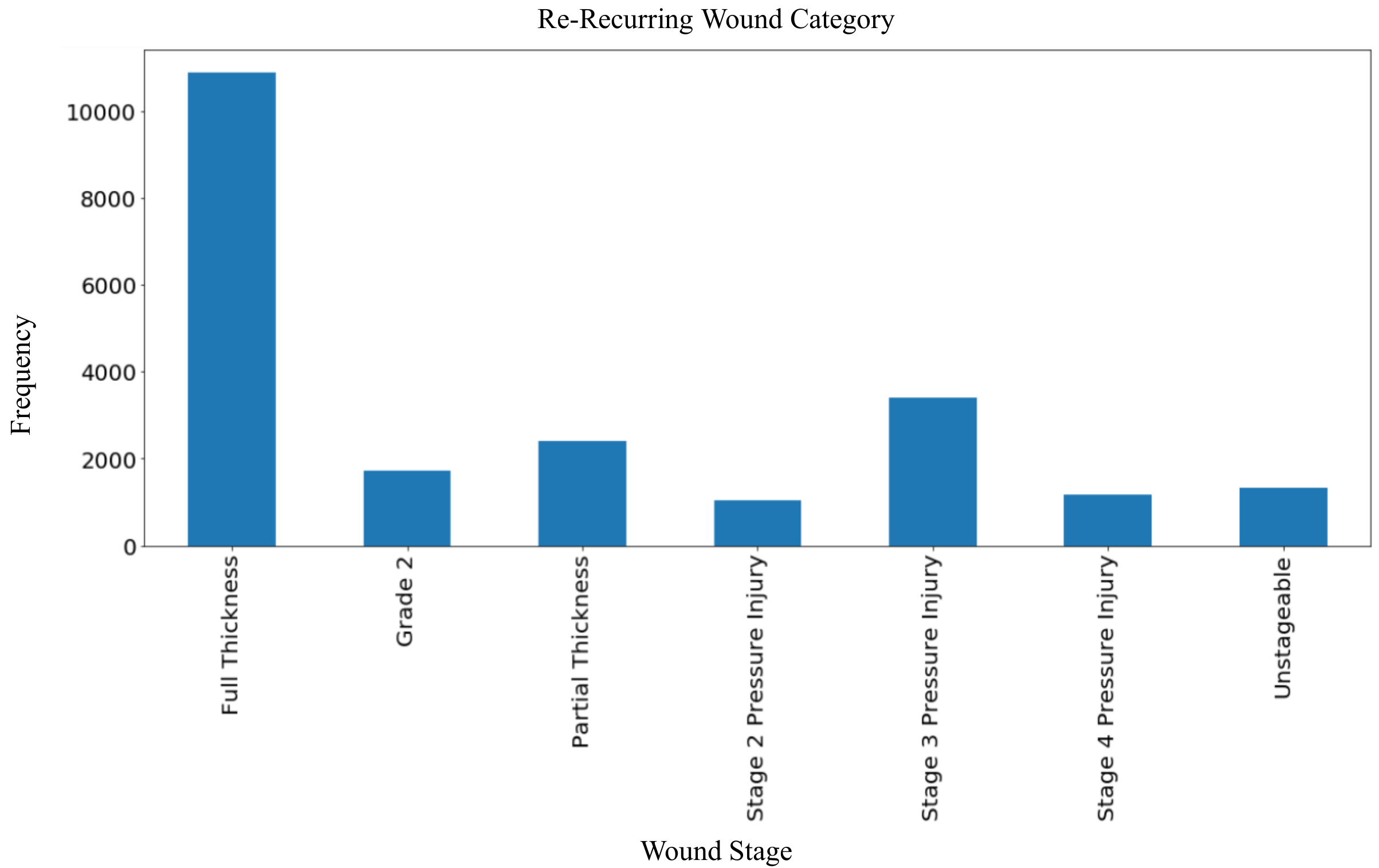}
    \caption{Wound Stage Histogram in Recurring Wound Class}
    \label{fig:stagehist}
\end{figure}

\noindent{\textbf{Wound Stage}}
The wound stage attribute represents the current Wound state by specifying the degree of severity. As the wound ages with out healing, the severity of the wound increases. Most patients delay the treatment process in hope of self-recovery which leads to further deterioration of the wound. As stage gets to a higher level it becomes more difficult to treat a wound. Figure~\ref{fig:stagehist} shows that wound with full-thickness stage is having the high chances of recurring in future.

\subsection{Episode-level Feature Analysis}
Table~\ref{tab:dataStats} middle row showcases the attributes used in episode-level risk of readmission.
Out of 39 episode-level attributes, we majorly discuss about the important features such as Total Chronic wounds treated, Days in treatment \& presence of Co-morbid conditions.

\noindent{\textbf{Admission Date:}}
The admission date attribute represents the date when the patient started receiving Wound care.
In the current data set, we observed that we had patients whose wound care episodes ranged from 2002-01-01 to 2020-03-20. In order to simulate the current health care trends we decided to limit the patient data set by including patients whose first Wound care episode was after the admission date of $\ge$ 2015-01-01.

\begin{figure}[t]
\centering
\includegraphics[width=\linewidth]{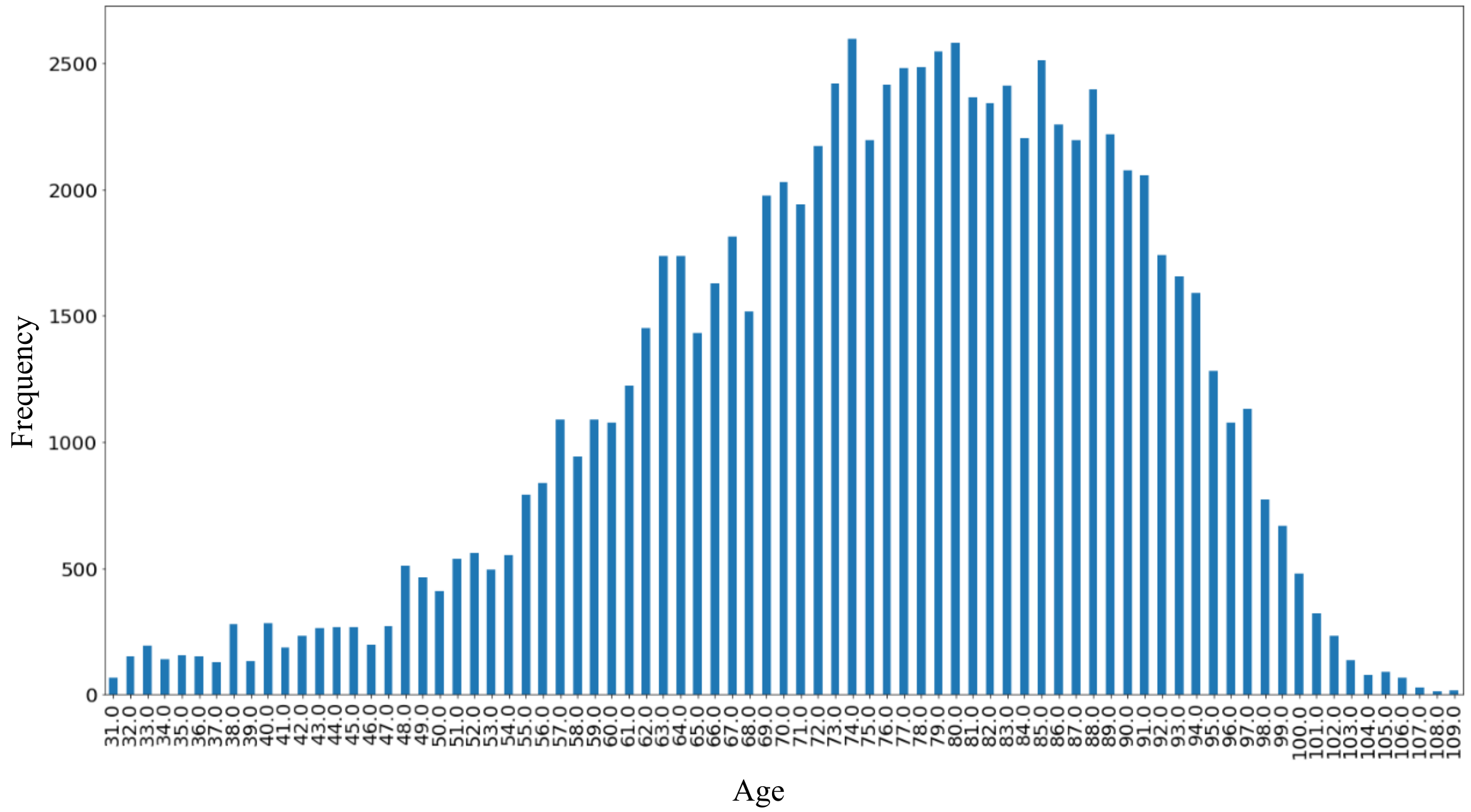}
    \caption{Patient Age Histogram}
    \label{fig:agehist}
\end{figure}

\noindent{\textbf{Patient Age:}}
Figure~\ref{fig:agehist} displays the patient age histogram for the combined data set where majority of patients were in the age range of 65 to 80.
The average patient age in the selected data set is 75, the maximum age is 110, and the minimum age is of 13 in our data set. As discussed earlier, age plays a critical role in formation of new wounds due to various medical complications that come with age.

\begin{figure}[t]
\centering
\includegraphics[width=\linewidth]{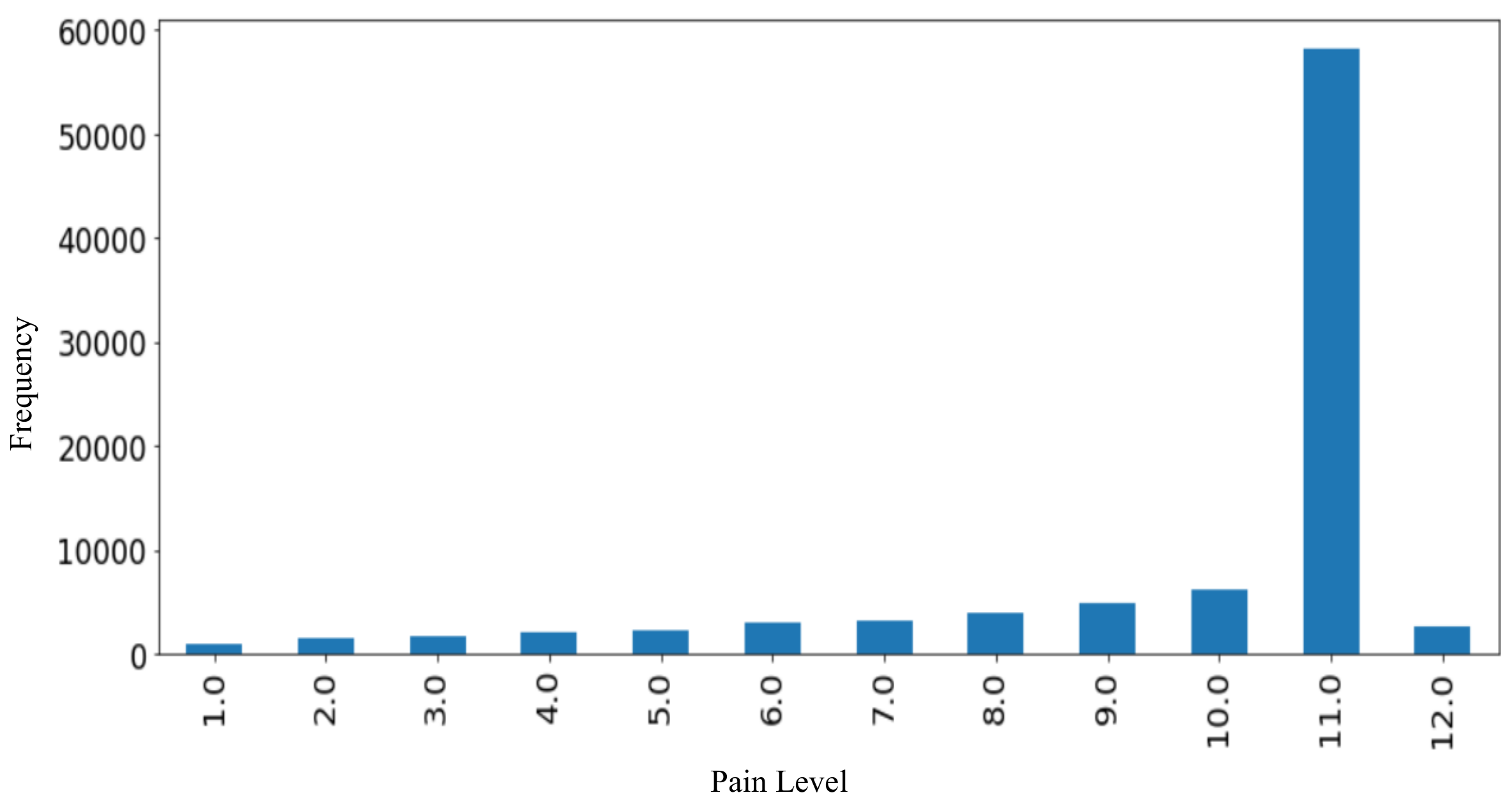}
    \caption{Pain-Level Histogram}
    \label{fig:painhist}
\end{figure}

\noindent{\textbf{Pain-Level}}
The pain-level of a wound describes the intensity of the pain experienced by the patient during the treatment period.
Figure~\ref{fig:painhist} shows the average pain scales of different patients during the treatment process. Observation from the Figure~\ref{fig:painhist} shows that majority of times a patient's pain-level is 11 which can be a primary driver of re-admission.

\noindent{\textbf{Total Chronic Wounds Treated}}
This attribute represents the count of chronic wounds such as diabetic, pressure, venous \& arterial ulcer that were treated during the episode. The higher number of chronic wounds not only represent a difficult road to healing but also represent high probability of patient re-admission.

\begin{figure}[t]
\centering
\includegraphics[width=\linewidth]{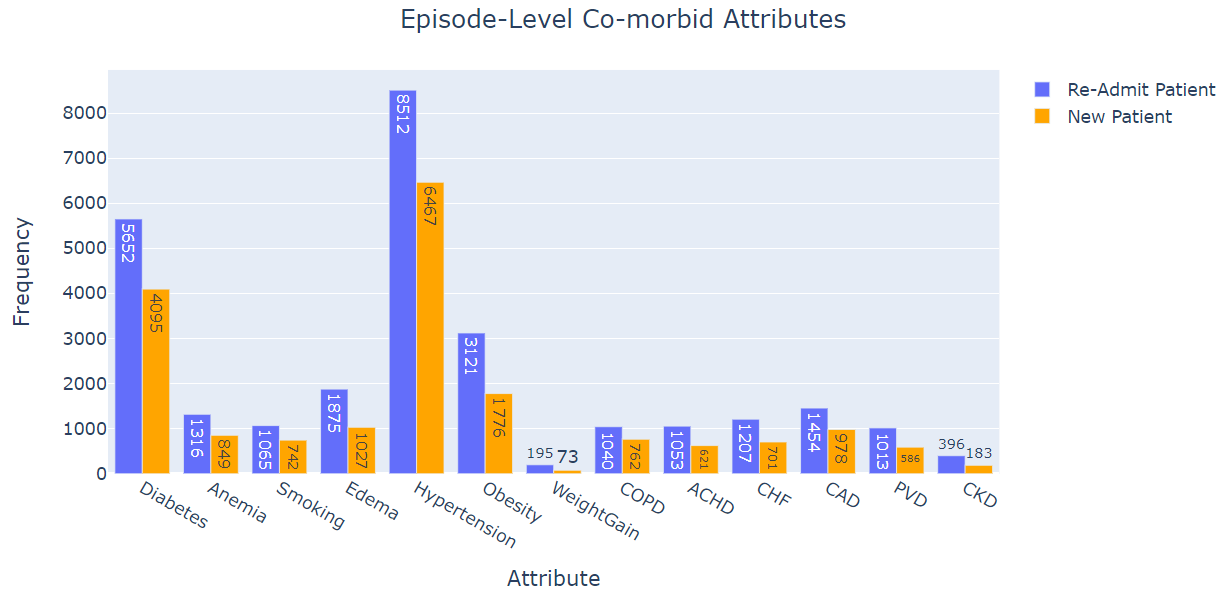}
    \caption{Episode-Level Co-morbid Attributes occurrence in New Patient vs Re-Admit Patient}
    \label{fig:comorbid}
\end{figure}

\noindent{\textbf{Presence of Co-morbid conditions}}
We have included multiple attributes that represent various co-morbid conditions such as Diabetes, Anemia, Chronic Obstructive Pulmonary Disease (COPD), \& Atherosclerotic Heart disease (AHD) that a patient might be having on them during the treatment cycle. The presence of co-morbid conditions is another indicator of high risk of re-admission for patients as these conditions deter healing \& also cause new wounds as shown in Figure~\ref{fig:comorbid}.

\section{Approach}
In practice, it is an arduous task to select the right imputation method for handling missing values \& it is hard to select the best classifier, and fine-tune it to specific parameters. 
Moreover, running each model manually on a large number of samples is a cost \& time-consuming task.
Current existing framework AutoML~\cite{feurer2015efficient} does not support imputation and calibration stages are particularly important for clinical prognostic modeling.
So, we employed a recent successful model AutoPrognosis useful for automated clinical prognostic models.
In this paper, we integrated some of the recently acclaimed tree-based classifiers LightGBM~\cite{ke2017lightgbm} and CatBoosting~\cite{prokhorenkova2018catboost} with AutoPrognosis framework.

\begin{figure}[t]
\centering
\includegraphics[width=\linewidth]{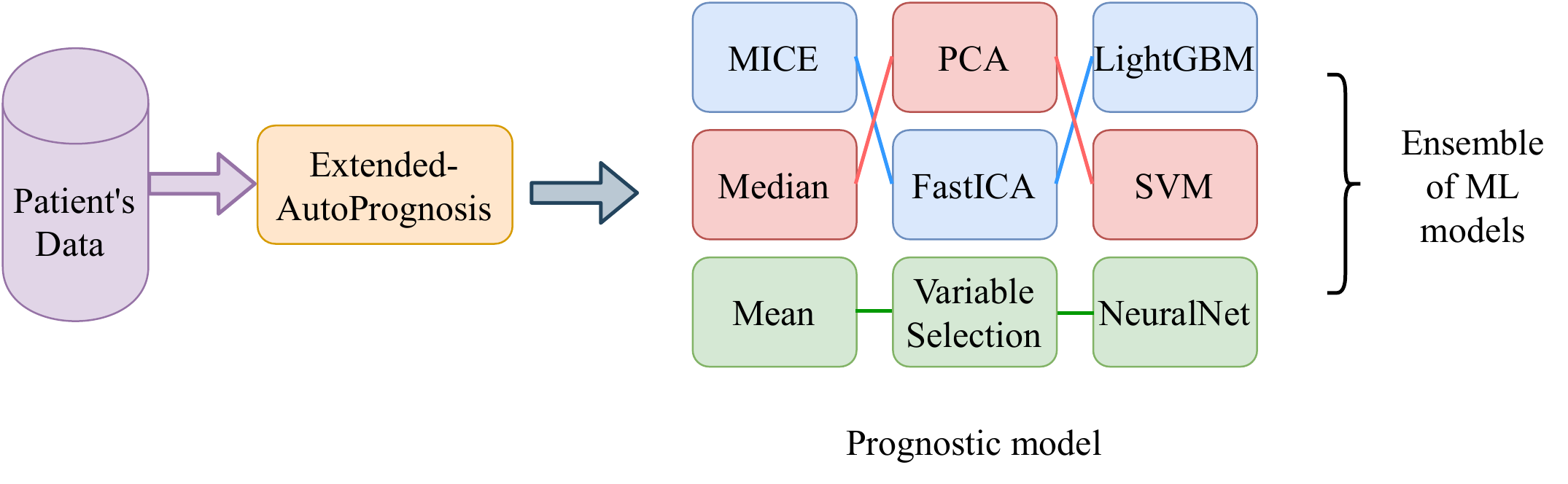}
    \caption{Extended-AutoPrognosis Model}
    \label{fig:model}
\end{figure}

\subsection{AutoPrognosis Framework}
The latest successful AutoPrognosis framework contains the following components, including,(i) 8 imputation algorithms, (ii) 10 feature processing techniques, (iii) 20 machine learning classifiers, (iv) 3 calibration methods, and total number of hyper-parameters (106), which is less than Auto-Sklearn framework (110)~\cite{feurer2019auto}. 
The core idea behind AutoPrognosis framework is to configures ML pipelines automatically, where every
pipeline comprises of 4 components mentioned above.
The Autoprognoisis pipeline configuration shown in Figure~\ref{fig:model} is as follows.
Let P=($A_{d}, A_{f}, A_{p}, A_{c}$) be a pipeline with the sets of imputation algorithms ($A_{d}$), feature processing algorithms ($A_{f}$), prediction algorithms ($A_{p}$), and calibration algorithms ($A_{c})$.  
The space of hyper-parameter configurations for a pipeline is $\Theta$ = $\Theta_{d}x\Theta_{f}x\Theta_{p}x\Theta_{c}$, where $\Theta_{d}$ being the hyper-parameter that corresponds to imputation algorithms $A_{d}$, and similarly for $\Theta_{f}$, $\Theta_{p}$, and $\Theta_{c}$.
Thus, the space of all possible pipeline configuration considered as $P_{\Theta}$, where $P_{\theta} \in P_{\Theta}$ is a selection of algorithms $p \in P$, and hyper-parameter settings $\theta \in \Theta$.

For a given clinical data (D), the main objective the AutoPrognosis framework is to find the best pipeline configuration $p_{\theta}^{*} \in P_{\Theta}$ is as follows

\begin{equation*}
    p_{\theta}^{*} \in \text{arg max}_{p_{\theta} \in P_{\Theta}} \frac{1}{K} \sum_{i=1}^{K} L(p_{\theta}; D_{train}^{i}, D_{valid}^{i}) 
\end{equation*}

Where $D_{train}^{i}$ and $D_{valid}^{i}$ are train and validation splits, L be the accuracy metric (macro avg precision, recall, etc), and i denotes the fold.

\section{Experimental Setup \& Results}

\begin{table*}[t]
\small
\centering
\caption{Wound-Level Results: Wound recurrence accuracy comparison for Extended-AutoPrognosis method, baseline Logistic Regression, LightGBM, and AutoPrognosis method}
\label{tab:woundlevel-rec}
\begin{tabular}{|l|c c c|c c c|c c c|c c c|}
\hline
\multicolumn{1}{|c|}{}&\multicolumn{3}{c|}{LR}&\multicolumn{3}{c|}{LightGBM}&\multicolumn{3}{c|}{AutoPrognosis} &\multicolumn{3}{c|}{AutoPrognosis-New}\\
\hline
Class$\downarrow$&P&R&F1&P&R&F1&P&R&F1&P&R&F1\\
\hline
\hline
Recurring Wound&0.66&0.28&0.39&0.77&0.61&0.67&0.76&0.64&0.69&0.76&0.66&0.70\\
\hline
New Wound&0.78&0.95&0.86&0.86&0.94&0.90&0.79&0.87&0.83&0.80&0.89&0.85\\
\hline
\end{tabular}

\end{table*}

\begin{table*}[t]
\small
\centering
\caption{Wound-Level Results: Patient Re-Admit accuracy comparison for Extended-AutoPrognosis method, baseline Logistic Regression, LightGBM, and AutoPrognosis method}
\label{tab:woundlevel-pc}
\begin{tabular}{|l|c c c|c c c|c c c|c c c|}
\hline
\multicolumn{1}{|c|}{}&\multicolumn{3}{c|}{LR}&\multicolumn{3}{c|}{LightGBM}&\multicolumn{3}{c|}{AutoPrognosis} &\multicolumn{3}{c|}{Extended-AutoPrognosis}\\
\hline
Class$\downarrow$&P&R&F1&P&R&F1&P&R&F1&P&R&F1\\
\hline
\hline
Re-Admit Patient&0.90&0.72&0.80&0.96&0.77&0.86&0.91&0.79&0.86&0.94&0.81&0.88\\
\hline
New Patient&0.77&0.92&0.84&0.81&0.97&0.88&0.85&0.96&0.88&0.86&0.98&0.90\\
\hline
\end{tabular}

\end{table*}

To reduce clinician workload, our first goal is to predict the patient's re-admission risk automatically from the selected attributes using three models trained separately on Wound level, Episode level, and  combined Wound Episode level data sets.
Using the these three data sets, our second goal is to build the weeks to readmit model.
We use 70:10:20 split for the train:validation:test for all our experiments.  

\noindent{\textbf{Dataset Details}}
Our patients re-admission risk data set is an accumulation of 5 years(Jan 2015–Jun 2020) of patients’ wound care data captured by a Wound care organization.   
Data collection was carefully done by following the survival model conditions to ensure that we cover the “Patient Demographics details”, “Procedures”,  “Medications”,  and  “Laboratory/Diagnosis of  Wound  condition”.   
Table~\ref{tab:dataStats} showcases the data set details by wound-level attributes, episode-level attributes, and the common attributes used across both the data sets.   
Also, we have two target columns present in wound-level data sets such as, wound-recurrence, and patient category.

\noindent{\textbf{Evaluation Metrics}}
We use classification metrics such as macro-average precision, recall, and F1-score to evaluate our methods.
To understand how each class is performing, we choose macro averaging that gives each class equal weight to evaluate systems performance across both two-classes.

For the second task (weeks to readmit), we use the standard error metrics such as, mean absolute error (MAE) and $R^2$-score to measure the model performance.

\subsection{Experiments on Wound-Level Data}
As part of first step, we use the 23 features to predict the Wound-Recurrence as well as patient category classification.
Table~\ref{tab:woundlevel-rec}, and~\ref{tab:woundlevel-pc} display the 5-fold cross-validation precision, recall, and F1-score results obtained using baseline logistic regression, LightGBM, AutoPrognosis, and Extended AutoPrognosis.

\noindent{\textbf{Wound-Recurrence Classification}}
Table~\ref{tab:woundlevel-rec} report the recurring wound classification results for wound-level data set.
Of the four training methods, we achieved best performance with extended Autoprognosis and worst with logistic regression.
We can observe that the addition of two classifiers LightGBM and CatBoost models improves the recall and F1-score for recurring wound class.
Further, we display the features that shows significant impact on improving the accuracy of the model in Table~\ref{tab:featureImp}.

\noindent{\textbf{Patients Category Classification}}
The target label recurring wound is available only on wound-level data set, but patients category column is available on both wound and episode data sets. 
To generalize the model, we predict the patients category (Re-Admitted or New Patient) using the four training methods mentioned above.
Table~\ref{tab:woundlevel-pc} describes the wound-level  patients category results where we achieved best performance with extended Autoprognosis and worst with logistic regression.
Moreover, we can observe that an increasing performance of recall and F1-score for both the classes when compared to recurring wound model.

\subsection{Experiments on Episode-Level Data}
The episode-level (sequence of patients visits) data set provides patients information at episode level rather than wound level.
Using the episode-level data set, our goal is to predict the patients category classification (Re-admit or New Patient) by considering the overall episodes information.
Similar to wound-level, we use all the four training methods to obtain the results on episode-level data set.

Table~\ref{tab:episodelevel-pc} shows results of precision, recall, and F1-score obtained on episode-level dataset consists of 39 features using the four methods.
Observation from the Tabe~\ref{tab:episodelevel-pc} that Extended-AutoPrognosis yields best performance compared to all the methods as well wound-level results and worst performance obtained from baseline logistic regression. 

\begin{table*}[!thb]
\small
\centering
\caption{Episode-Level Results: Patient Category accuracy comparison for Extended-AutoPrognosis method, baseline Logistic Regression, LightGBM, and AutoPrognosis method}
\label{tab:episodelevel-pc}
\begin{tabular}{|l|c c c|c c c|c c c|c c c|}
\hline
\multicolumn{1}{|c|}{}&\multicolumn{3}{c|}{LR}&\multicolumn{3}{c|}{LightGBM}&\multicolumn{3}{c|}{AutoPrognosis} &\multicolumn{3}{c|}{Extended-AutoPrognosis}\\
\hline
Class$\downarrow$&P&R&F1&P&R&F1&P&R&F1&P&R&F1\\
\hline
\hline
Re-Admit Patient&0.95&0.69&0.80&0.93&0.75&0.83&0.94&0.76&0.85&0.90&0.79&0.86\\
\hline
New Patient&0.78&0.97&0.86&0.83&0.95&0.88&0.83&0.97&0.90&0.84&0.98&0.91\\
\hline
\end{tabular}

\end{table*}

\begin{table*}[!thb]
\small
\centering
\caption{WoundEpisode-Level Results: Patient Category accuracy comparison for Extended-AutoPrognosis method, baseline Logistic Regression, LightGBM, and AutoPrognosis method}
\label{tab:woundepisodelevel-pc}
\begin{tabular}{|l|c c c|c c c|c c c|c c c|}
\hline
\multicolumn{1}{|c|}{}&\multicolumn{3}{c|}{LR}&\multicolumn{3}{c|}{LightGBM}&\multicolumn{3}{c|}{AutoPrognosis} &\multicolumn{3}{c|}{Extended-AutoPrognosis}\\
\hline
Class$\downarrow$&P&R&F1&P&R&F1&P&R&F1&P&R&F1\\
\hline
\hline
Re-Admit Patient&0.89&0.64&0.74&0.98&0.87&0.92&0.92&0.91&0.92&0.92&0.92&0.92\\
\hline
New Patient&0.87&0.97&0.92&0.87&0.98&0.93&0.90&0.97&0.94&0.91&0.98&0.94\\
\hline
\end{tabular}

\end{table*}

\begin{table}%
\small
\centering
\caption{Weeks to Re-Admit prediction comparison for LightGBM method and the baseline Linear Regression on MAE. LR=Linear Regression.}
\label{tab:weeks2readmit}
\begin{tabular}{|l|c|c|}
\hline
Feature set$\downarrow$&\multicolumn{1}{c|}{LR}&\multicolumn{1}{c|}{LGBM}\\
\hline
\hline
Wound-Level&3.2&2.6\\
\hline
Episode-Level&4.6&3.1\\
\hline
WoundEpisode-Level & 3.0 & 2.3\\
\hline
\end{tabular}

\end{table}

\subsection{Experiments on WoundEpisode-Level Data}
Here, we combine the wound and episode level datasets by using the common patient id across both the datasets resulted in overall 66 attributes. 
The common attributes across two datasets reported in Table~\ref{tab:dataStats}.
Using the woundepisode-level dataset, our final goal is predict the patients category classification (Re-admit or New Patient) by considering the first episode, or last episode, or overall episodes information.

Table~\ref{tab:woundepisodelevel-pc} report results of precision, recall, and F1-score obtained on woundepisode-level dataset using the four methods.
Observation from the Table~\ref{tab:woundepisodelevel-pc} that Extended-AutoPrognosis yields best performance compared to all the methods as well wound and episode level results and worst performance obtained from baseline logistic regression. 
We also observe that the woundepisode based data results were better than individual datasets when we use Extended AutoPrognosis  model,  with  0.95  recall  for  the Re-Admitted patient  class.  
Overall, we observe that our Extended AutoPrognosis models provide the best results.

\subsection{Weeks to Re-Admit Prediction} For the weeks to readmit prediction our labeled data follows a power law distribution. 
To perform the weeks to readmit prediction, we use the same features as used in the patient category classification model.
However, here the target variable is the weeks to readmit for a particular wound-level, episode-level, and wound episode-level data set. 
The results in Table~\ref{tab:weeks2readmit} illustrate the performance of the LightGBM model in comparison with the baseline linear regression model. 
To measure the model performance, we use mean absolute error (MAE) as the metric.
The minimum number of weeks to readmit is one, the maximum is 15 weeks, and the average is 6 weeks in our dataset. 
The LightGBM model achieves an MAE of 2.6 weeks if we consider only wound-level features, 3.1 MAE on episode-level features, and 2.3 MAE on woundepisode-level features.

\subsection{Feature Importance Analysis}
Table~\ref{tab:featureImp} display the significant impact of features across wound and episode level data sets.
We observed that age and BMI are the most important patient attributes.  
Wound Location, Type, Stage were the best predictor for the ReAdmit/New Patient classifier, which is expected since the wound attributes intuitively correlates with the seriousness of the wound.  
Similarly, Non-Compliance attributes plays a major role in risk of readmit prediction. 

\begin{table}%
\centering
\small
\caption{Feature Importance Analysis (Task 1: Wound-level Patient Category Risk prediction; Task 2: Episode-level Patient Category Risk prediction). Features are listed in descending order of importance}
\label{tab:featureImp}
\resizebox{0.48\textwidth}{!}{\begin{tabular}{|l|l|l|}
\hline
No&Task 1&Task 2\\
\hline
\hline
1&Days in TX for Patients&Avg BMI\\
\hline
2&Patient Age&Avg Days in TX for Wounds\\
\hline
3&Days Prior to TX&Patient Age\\
\hline
4&Days in TX for Wounds&Patient Discharge Status\\
\hline
5&Wound Location&Avg Pain Level for Episode\\
\hline
6&Patient Discharge Status&NonCompliance Wound Visits Rate\\
\hline
7&Wound Type&NonCompliance Dressing Rate\\
\hline
8&Visits for Wound&NonCompliance OffLoad Rate\\
\hline
9&Avg Pain Level for Wound&NonCompliance Limb Rate\\
\hline
10&Wound Stage&Lower Extremity Wounds for Episode\\
\hline
11&NonCompliance Dressing Rate&Chronic Wounds for Episode\\
\hline
12&NonCompliance Wound Visits Rate&NonCompliance Diet Rate\\
\hline
13&NonCompliance OffLoad Rate&Wounds for Episode\\
\hline
14&NonCompliance Limb Rate&Edema\\
\hline
\end{tabular}}

\end{table}

\section{Conclusion}
In this paper, we present the
patient’s risk of readmission model using extended-autoprognosis method. 
To achieve the first goal, our proposed system, mainly depends on three data sets such as, (i) would-level information, (ii) episode-level information, and (iii) combination
of wound and episode level information. 
Our second goal is to predict the time period during which a readmission event, i.e weeks from the date of discharge, can be expected.  
We are the first to perform extensive experiments on a large dataset for patients risk of readmit on wound care data.  
Our experiments show that extended-autoprognosis offers accurate re-admission predictions, and therefore can be practically be deployed.  
This model can serve as a useful tool for care managers to have better insight about a patient's re-admission and preemptively prevent avoidable re-admissions.

\bibliographystyle{ACM-Reference-Format}
\bibliography{sample-base}

\appendix

\end{document}